\documentclass[runningheads]{llncs}
\usepackage{graphicx}
\usepackage{amsmath,amssymb} 
\usepackage{xcolor}
\usepackage{booktabs}
\usepackage[hyphens]{url}
\usepackage{multirow}
\usepackage{subcaption}
\usepackage{graphics}
\usepackage{graphicx}

\usepackage[width=122mm,left=12mm,paperwidth=146mm,height=193mm,top=12mm,paperheight=217mm]{geometry}

\begin{document}
\pagestyle{headings}
\mainmatter
\def\ECCV16SubNumber{3755}  


\title{Learning to Censor by Noisy Sampling}

\titlerunning{Learning to Censor by Noisy Sampling}

\author{Ayush Chopra\inst{1} \and
Abhinav Java\inst{2}\and
Abhishek Singh\inst{1}\and
Vivek Sharma\inst{1}\and
Ramesh Raskar\inst{1}
}
\authorrunning{Chopra et al.}
%
\institute{Massachusetts Institute of Technology, Cambridge, MA \and
Delhi Technological University, Delhi, India \\}

\maketitle

\begin{abstract}
Point clouds are an increasingly ubiquitous input modality and the raw signal can be efficiently processed with recent progress in deep learning. This signal may, often inadvertently, capture sensitive information that can leak semantic and geometric properties of the scene which the data owner does not want to share. The goal of this work is to protect sensitive information when learning from point clouds; by censoring the sensitive information before the point cloud is released for downstream tasks. Specifically, we focus on preserving utility for perception tasks while mitigating attribute leakage attacks. The key motivating insight is to leverage the localized saliency of perception tasks on point clouds to provide good privacy-utility trade-offs. We realise this through a mechanism called \textit{Censoring by Noisy Sampling} (\textit{CBNS}), which is composed of two modules: i) Invariant Sampler: a differentiable point-cloud sampler which learns to remove points invariant to utility and ii) Noisy Distorter: which learns to distort sampled points to decouple the sensitive information from utility, and mitigate privacy leakage. We validate the effectiveness of CBNS through extensive comparisons with state-of-the-art baselines and sensitivity analyses of key design choices. Results show that CBNS achieves superior privacy-utility trade-offs on multiple datasets.

\end{abstract}


\section{Introduction}

Proliferation of 3D acquisition systems such as LiDARs, ToF cameras, structured-light scanners has made it possible to sense and capture the real-world with high fidelity. Point clouds are emerging as the preferred mode to store the outputs of these 3D sensors given that they are lightweight in memory and simple in form. Recent advances in deep learning have allowed to directly process the raw sensor output; which has enabled use of point clouds for diverse perception tasks across classification~\cite{qi2017pointnet,qi2017pointnet++,dgcnn,li2018pointcnn,wu2019pointconv,zhao2021_point_transformer}, semantic segmentation~\cite{xu2020squeezesegv3,hu2020randla,zhang2020polarnet,chen2020hapgn}, object detection~\cite{qi2018frustum,qi2019deep,martin2018complex,shi2019pointrcnn}, and registration~\cite{register-1,register-2}. This is facilitating algorithms for critical applications across autonomous navigation, precision surgery and secure authentication.

The deployment of downstream algorithms in these critical domains implies that the sensor often captures sensitive information, which the user would like to keep private. This is then inadvertently encoded in representations learned from the signal~\cite{song2019overlearning}, leaking several semantic and geometric properties of the scene. Consider for instance, the robotic vacuum cleaners which use LiDAR sensors to efficiently navigate inside the house. The captured signal is also sufficient to localize and map the entire house (via SLAM) as well as track and surveil individuals (via object detection). Similarly, this is also valid for the popular \textit{face-id} experience in recent smartphones which use structured light to capture point clouds of the owner(s) face and use it for authentication, locally on-device. It is well understood that a lot of semantic information (age, gender, expression etc.) can be perceived from the point cloud - which the user may not be willing to share. With the emergence of strict regulations on data capture and sharing such as HIPAA~\cite{hippa}, CCPA~\footnote{\url{https://leginfo.legislature.ca.gov/faces/billTextClient.xhtml?bill_id=201720180AB375}}, capturing such sensitive information can create legal liabilities. The goal of this paper is to alleviate such privacy concerns, while preserving utility, by transforming the point cloud to censor sensitive information \textit{before} it is released for downstream utility tasks.

In practice, the design of such transformation functions depends upon the definition of the utility task and privacy attack. Most prior work has focused on preserving the utility of geometric tasks (image-based localization, SLAM, SfM, etc.) while protecting against input reconstruction attacks~\cite{DBLP:journals/corr/AroraLM15}. For these setups, the  dominant idea is to transform the point cloud into 3D line cloud~\cite{linecloud-ImageBased} which obfuscates the semantic structure of the scene while preserving utility for camera localization~\cite{linecloud-Camera}, SLAM~\cite{linecloud-SLAM}, SfM~\cite{linecloud-SFM} etc. In contrast, we focus on providing utility for perception tasks (classification, detection, segmentation etc.) while mitigating sensitive attribute leakage~\cite{jia2018attriguard}. We posit that projecting to line clouds is an infeasible transformation for perception tasks because: i) line clouds disintegrates the semantic structure of the scene required for perception which worsens the utility. This is visualized in~\cite{linecloud-ImageBased} and validated by our analysis in section~\ref{sec:discussion}; and ii) line clouds are now also vulnerable to inversion attacks, as recently shown in ~\cite{chelani2021privacypreserving}, which worsens the privacy. We propose \textit{Censoring by Noisy Sampling (CBNS)} as an alternate transformation for censoring point clouds, which provides improved privacy-utility trade-offs.

The motivating insight for \textit{CBNS} is that performance on perception tasks (utility) only depends upon only a small subset of points (critical points) such that removing (or \textit{sampling}) other non-critical points does not change prediction. Leveraging this for censoring point clouds presents two challenges: \textit{First}, conventional point cloud sampling methods are designed to improve compute efficiency while retaining maximal information about a specific task. Hence, we need to design methods that can jointly sample critical points for the utility task and remove information \textit{invariant} to utility. \textit{Second}, this invariant sampling is necessary but not sufficient, as critical points for task and sensitive attributes can overlap; as we observe through quantitative analysis in section~\ref{sec:premise-val}. We develop \textit{CBNS} to overcome these challenges - i) by introducing an invariant sampler that balances privacy-utility trade-off in its sampling via an adversarial contrastive objective~($\ell_{aco}$); ii) by designing a noisy distortion network that adds sample-specific noise to minimize the overlap between task and sensitive information in an utility conducive manner. We demonstrate the effectiveness of our solution in section~\ref{sec:expts}.

\textbf{Contributions: } Our CBNS is an end-to-end learning framework for protecting sensitive information in perception tasks by dynamically censoring point clouds. CBNS is composed of: i) an invariant sampler that learns to sample non-sensitive points in a task-oriented manner by balancing privacy-utility, ii) a noisy distorter that learns to randomize sampled points for utility conducive removal of sensitive information. We demonstrate the effectiveness of our framework through extensive comparisons against strong baselines and analyses of key design choices. Results show that CBNS significantly improves privacy-utility trade-offs on multiple datasets.


\section{Problem Formulation}
This section formalises the notation for our task, the threat and attack models, and our privacy definition. 
\label{sec:formulation}

\noindent \textbf{Notation:} Consider a data owner $O$ with a point cloud dataset $D_{O} = (P, Y)$ of $N$ datapoints and ($p, y$) denotes a paired sample, for $p \in P$ and $y \in Y$. Specifically, $p \in R^{m \times d}$ is a \textit{point cloud} defined as an unordered set of $m$ elements with $d$ features; and $y$ is a \textit{label set} of $k$ attributes describing $p$. For instance, $p$ can be a 3D point cloud representing a human face ($p \in R^{m \times 3}$) with the \textit{set} $y$ containing categorical attributes that indicate the \textit{\{age, gender, expression\}} ($k=3$) of the individual. For every pair ($p, y) \in D_{O}$, certain attributes in the label set $y$ represent sensitive information which the data owner ($O$) wants to keep private ($y_s$) but is comfortable sharing the non-sensitive (or task) information ($y_{t}$), such that ($y=y_s \cup y_{t}$). The risk of leaking this sensitive information prevents the data owner from sharing $D_{O}$ with untrusted parties; especially with recent progress in deep learning where  attackers can efficiently learn functions ($F$) that can directly map the raw point cloud $p$ to any attribute $a \in y$, where $a = F(p)$~\cite{shi2019pointrcnn,zhao2021_point_transformer}. Trivially omitting $y_{s}$ from $y$ to share the dataset of paired samples \{($p, y_{t}$)\} is not enough since the sensitive information is encoded in $p$ and can be inferred by attackers (\textit{e.g.} using pre-trained models or auxilliary datasets). Hence, to facilitate data sharing with untrusted parties, it is essential to \textit{censor} the sensitive information in $p$ (that leaks $y_{s}$) \textit{before} the dataset can be released. Our goal is to learn such a transformation function ($T(\theta_T;\cdot)$) that censors each sample in $D_{O}$ by generating $\hat{p} = T(p)$. This allows to release ($\hat{p}, y_{t}$) instead of ($p, y_{t}$). Henceforth, we denote this \textit{censored dataset} as $(\hat{P}, \hat{Y})$. The key challenge for $T$ is to preserve utility of the task information ($y_{t}$) while protecting privacy of sensitive information ($y_{s}$). In practice, the design of $T$ depends upon definition of the utility task and privacy attack. We focus on providing utility for perception tasks while mitigating attribute leakage attacks~\cite{jia2018attriguard}.

\begin{figure}[t]
    \centering
    \includegraphics[width=0.95\textwidth]{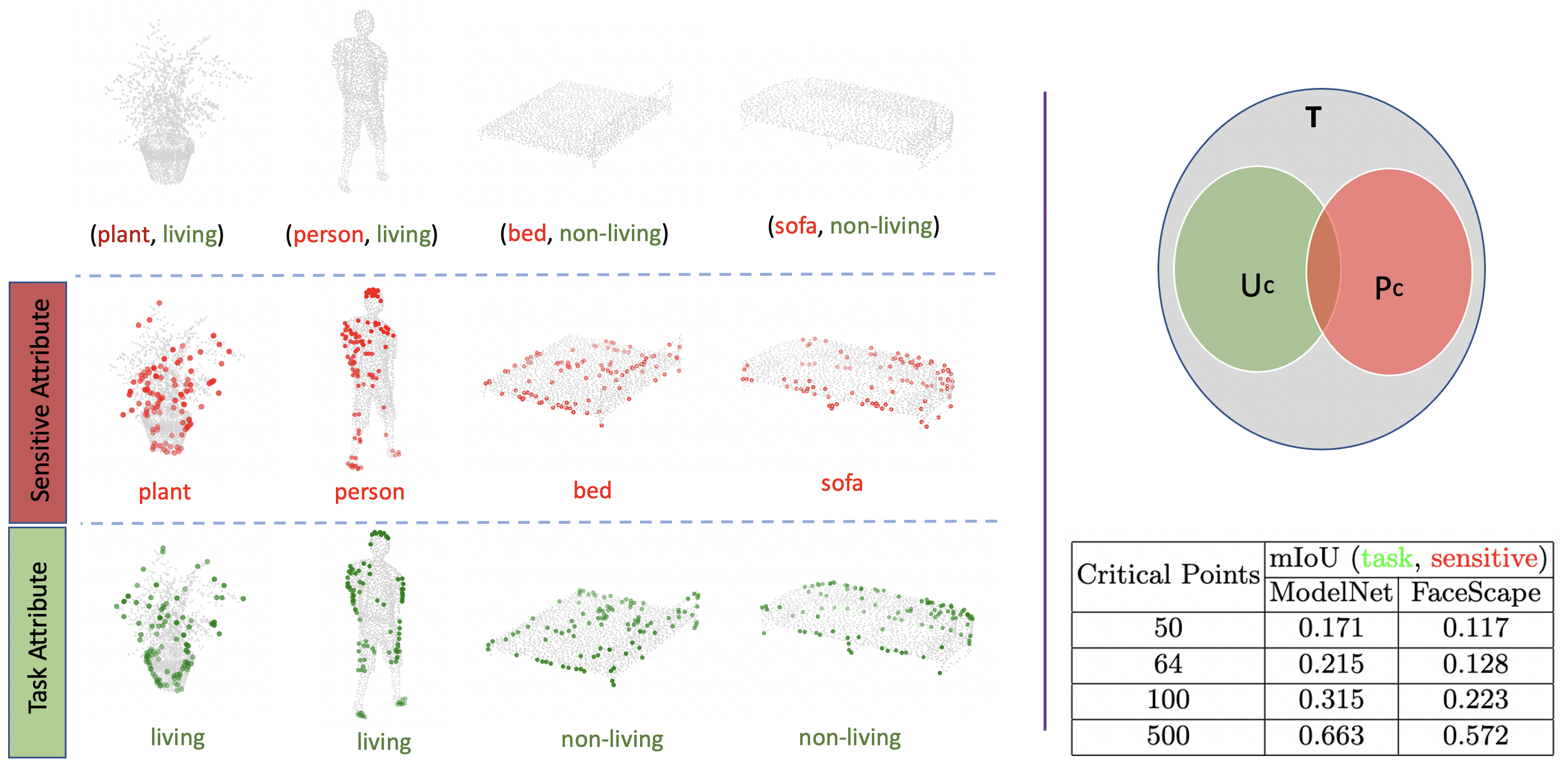}
    \caption{\textbf{Premise Validation}. (\textit{Left}) - perception on point clouds depends upon few critical points. (\textit{Right - bottom}) table shows overlap of critical points for sensitive ($P_c$) and task ($U_c$) attributes; (\textit{Right - top}) goal of a censoring mechanism is to remove $\textcolor{gray}{T} - \textcolor{green}{U_c}$ and reduce $\textcolor{red}{P_c} \cap \textcolor{green}{U_c}$. We bridge these ideas in section~\ref{sec:premise-val} and introduce CBNS in section~\ref{sec:method} to accomplish both goals.}
    \label{fig:premise-validation}
\end{figure}

\noindent \textbf{Threat Model:} We assume that the attacker gains access to a subset of censored point clouds ($\hat{P}, Y$) intending to infer sensitive attributes ($y_s$). This is practical since data owners typically share data with external entities for storage and also for monetary incentives. Further, the threat is also valid if the attacker gains access to the learned censoring function $T$ which can be used to simulate a dataset that mimics the censored point cloud distribution. This is practical if the attacker is one of the data owners that has access to $T$. We note that unlike differential privacy~\cite{dwork2006calibrating} that protects identifiability, our threat model protects sensitive attribute leakage~\cite{jia2018attriguard}.

\noindent \textbf{Attack Model:} We model an attacker that uses the released dataset to train state-of-the-art DNN models that can directly predict the sensitive attribute from the point cloud. This attacker may use arbitrary models which are not accessible during training, and hence we mimic a proxy attacker for learning the censoring transformation. We represent the proxy attacker by a state-of-the-art DNN parameterized as $f_{A}(\theta_{A};\cdot)$ and trained on censored point clouds $\hat{P}$.




\noindent \textbf{Privacy:} Following the setup described by Hamm~\textit{et al.}~\cite{JMLR:v18:16-501}, we define privacy as the expected loss over the estimation of sensitive information by the attacker. This privacy loss $L_{priv}$, given $\ell_p$ norm, for an attacker can be stated as:
$$L_{priv}(\theta_T,\theta_{A})\triangleq E[\ell_p(f_{A}(T(p;\theta_T);\theta_{A}), y_{s})]$$
Under this definition, releasing sensitive information while preserving privacy manifests as a \textit{min-max} optimization between the data owner and the attacker. However, for training the model parameters, we use a proxy adversary from which gradients can be propagated. We refer to the attack performed by this proxy attacker as an \textit{online attack} and note that this allows mimicking worst-case setups where the attacker can also dynamically adapt using protected data and sensitive label information~\cite{singh2021disco}. 
We note that our definition of privacy significantly differs from differential privacy~\cite{dwork2006calibrating} since we aim to protect sensitive attributes instead of the identity of the data owner.

\section{Methodology}
In this section, we introduce \textit{Censoring by Noisy Sampling (CBNS)} - a mechanism to censor point clouds for enabling utility of perception tasks while protecting leakage attack on sensitive attributes. We begin by discussing our key motivating insight and then delineate the proposed \textit{CBNS} mechanism.

\subsection{Premise Validation}
\label{sec:premise-val}


State-of-the-art DNN models such as PointNet~\cite{qi2017pointnet}, PointNet++~\cite{qi2017pointnet++}, DGCNN~\cite{dgcnn} have successfully handled the irregularity of the raw point cloud and achieved remarkable progress on perception tasks such as classification, segmentation etc. Extensive empirical analysis of these networks shows that classification performance depends upon only a small subset of points (\textit{critical points}) such that removing other non-critical points does not change prediction. Figure~\ref{fig:premise-validation} visualizes the critical points for perceiving the category (\textit{plant, person, bed, sofa}) and super-type (\textit{living, non-living}) of a few ModelNet dataset samples~\cite{modelnet} by training PointNet. The observed \textit{localized} (i.e. depends on critical points) and \textit{task-oriented} (different across category and super-type) saliency is a key motivating insight for censoring point clouds for privacy-utility release.

Assume a privacy-utility scenario where the super-type is task (utility) and the category is the sensitive attribute (privacy). In principle, we achieve good utility (predicting super-type) by only keeping the necessary critical points. Since critical points are visualized via post-training analysis, in practice, this presents two challenges for data release: \textit{First,} conventional point cloud sampling methods are designed to improve compute efficiency while retaining maximal information for a specific task. Hence, we need to design methods that can jointly sample critical points for the utility task and remove information \textit{invariant} to utility. \textit{Second}, this invariant sampling is necessary but not sufficient, as critical points for task and sensitive attributes can overlap. For instance, the top-100 critical points for \textcolor{green}{super-type} and \textcolor{red}{category} in ModelNet have mIoU of \textit{31\%} (table in figure~\ref{fig:premise-validation}). Hence we also need to distort the sampled points to decouple the sensitive and task attributes. The venn-diagram in figure~\ref{fig:premise-validation}
helps visualize this constraint. Intuitively, we want to learn a censoring transformation that can concurrently remove $\textcolor{gray}{T} - \textcolor{green}{U_c}$ and reduce $\textcolor{red}{P_c} \cap \textcolor{green}{U_c}$. With this motivation, next we describe our proposed mechanism to censor point clouds.

\begin{figure}[t]
    \centering
    \includegraphics[width=\linewidth]{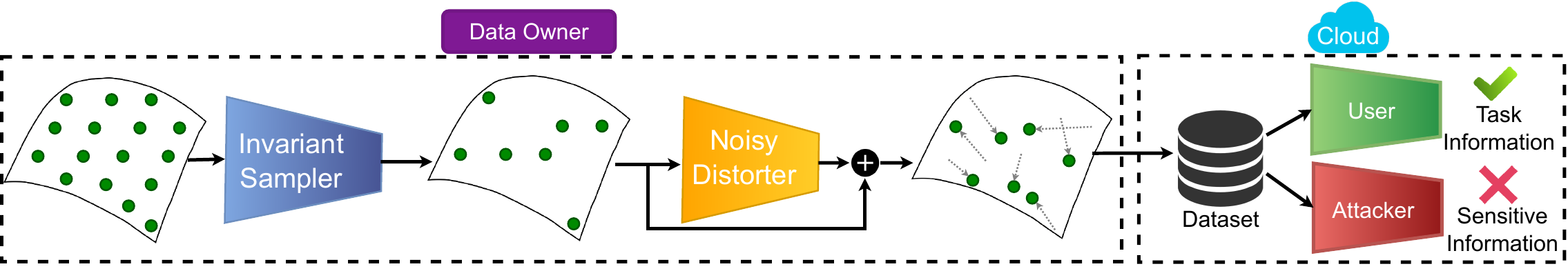}
    \caption{\textbf{Censoring by Noisy Sampling} through a a three-player game: i) Data Owner ($O$), ii) User ($U$) and iii) Attacker ($A$). $O$ censors every sample in the dataset and shares it with $U$ to train a model on the task information. $A$ intercepts the released dataset and attempts to leak the sensitive information. We design \textit{CBNS}, composed of two modules: a) \textit{Invariant Sampler} and b) \textit{Noise Distorter}, to help $O$ enable $U$'s task and avert $A$'s attack. The design of the mechanism is delineated in section~\ref{sec:method}.}
    \label{fig:censor-by-sample}
\end{figure}

\subsection{Censoring by Noisy Sampling}
\label{sec:method}
The task of censoring to mitigate information leakage involves three key entities: i) Data Owner ($O$), ii) User ($U$) and iii) Attacker ($A$). $O$ censors each sample in the dataset to protect sensitive information and releases it for $U$, an untrusted but honest entity, to learn a model on the non-sensitive information. $A$ intercepts the released dataset and queries it to leak the sensitive information. We design \textit{Censoring by Noisy Sampling} (\textbf{CBNS}) to help $O$ facilitate the task of $U$ and prohibit the task of $A$. This three-player game~\cite{max-entropy} is summarized in figure~\ref{fig:censor-by-sample} and described below:

\textbf{\underline{a) Owner}} owns the point cloud dataset ($P, Y$) which is to be released. This entity censors the sensitive information in each sample ($p, y$) for $p \in P$ and $y \in Y$. CBNS is composed of two parametric modules, applied sequentially: i) \textit{Invariant Sampler} ($f_{S}(\theta_S; \cdot)$), ii) \textit{Noisy Distorter} ($f_{D}(\theta_{D}; \cdot) $). 

\textit{First}, $p \in R^{m \times d}$ is passed through $f_{S}(\theta_{S}; \cdot)$, differentiable DNN sampler built upon~\cite{lang2020samplenet}, which selects a subset of $r$ points relevant for encoding task information to generate an intermediate point cloud $p_{s} \in R^{r \times d}$, where $r << m$. In contrast to conventional sampling methods which are aimed at improving compute efficiency while preserving all information, $f_{S}$ is a lossy sampler designed to remove points \textit{invariant} to utility. The extent and quality of censoring depends upon design of $f_{S}$, which we analyse in section~\ref{sec:discussion}. Releasing this $p_{s}$ may still leak sensitive attribute through points which overlap with utility (mIoU table in fig~\ref{fig:premise-validation}). \textit{Next}, $p_{s}$ is passed through $f_{D}(\theta_{D}; \cdot)$ which generates task-oriented noise to distort $p_{s}$. This is done to decouple overlapping sensitive (privacy) and task (utility) information and executed in the following steps: i) $\mu, \sigma = f_{D}(p_{s};\theta_D)$, ii) $z_{s} \sim \mathcal{N}(\mu, \, \sigma^{2})$, iii) $\hat{p} = p_{s} + z_{s}$ where the sampled noise $z_{s} \in R^{r \times d}$. All the censored point cloud samples $(\hat{p}, \hat{y})$ are aggregated into the dataset ($\hat{P}, \hat{Y}$) and then released for use by untrusted parties.

\textbf{\underline{b) User}} is an untrusted but honest entity that receives the released dataset $(\hat{P}, \hat{Y})$ and uses it to infer the non-sensitive task attributes. The \textit{User} trains a state-of-the-art DNN model that can learn to directly map the raw point cloud signal to the task attribute ($y_{t}$). For training \textit{CBNS}, we mimic the real user using a proxy user which is parameterized with $f_{U}(\theta_{U}; \cdot)$ that consumes $\hat{p} \in \hat{P}$ to predict $y_{t} \in \hat{Y}$.

\textbf{\underline{c) Attacker}} is an untrusted semi-honest entity that acquires access to ($\hat{P}, Y$) with the intention to leak sensitive information about the data owner. The attacker is parameterized with ($f_{A}(\theta_{A}; \cdot)$) which is not accessible during design of \textit{CBNS}. Hence, for training, we use a proxy attacker that consumes $\hat{p} \in \hat{P}$ to predict the sensitive attribute ($y_{s}$). We note that $f_{A}$ is a proxy attacker used for training \textit{CBNS}, while a \textit{distinct} \textit{offline attacker} ($f$), not used for training, is employed for evaluation tasks.

\subsubsection{Training:}
\label{sec:training}
The utility loss is approximated based on the performance of the proxy user which depends upon parameters $\theta_T$ ($ \theta_T = \theta_S \cup \theta_D$) and $\theta_{U}$ that are learned during training. The objective function is given by: 
\begin{equation}
\label{eq:util}
L_{util}(\theta_S, \theta_D, \theta_{U})\triangleq E[\ell_u(f_{U}(f_D(f_S(p;\theta_S)\theta_D);\theta_{U}), y_{t})] 
\end{equation}
where, $\theta_S, \theta_D$ are parameters of CBNS and $\theta_U$ are parameters of the the proxy user network; and $\ell_u$ is the cross entropy loss ($\ell_{cce}$).

The privacy loss is approximated based on the performance of the proxy attacker which depends upon parameters $\theta_T, \theta_A$ learned during training. The objective function is given by:
\begin{equation}
\label{eq:priv}
L_{priv}(\theta_S, \theta_D, \theta_A)\triangleq E[\ell_a(f_{A}(f_D(f_S(p;\theta_S)\theta_D);\theta_{A}), y_{s})] 
\end{equation}
where, $\theta_S, \theta_D$ are parameters of CBNS and $\theta_A$ are parameters of the proxy attacker; and $\ell_{a}$ denotes the attacker loss. For training CBNS, we define $\ell_{a}$ with the following objective function:
\begin{equation}
\label{eqn:attacker-loss}
    \ell_{a} = \alpha * (\ell_{cce}(f_A(f_D(p_s)), y_{s})) + (1 - \alpha)* \ell_{aco}(f_D(p_{s}), y_s , y_t)
\end{equation}
where $\ell_{cce}$ is categorical cross-entropy and $\ell_{aco}$ is an adversarial contrastive loss, inspired from~\cite{osia}; and $\alpha$ is a scalar hyperparameter.\\
\textbf{Adversarial Contrastive Loss ($\ell_{aco}$):} Our analysis in section~\ref{sec:discussion} shows that using $\ell_{aco}$ significantly improves privacy-utility trade-offs. Consider, for instance, age ($y_{s}$) to be the sensitive attribute. In the conventional contrastive loss, we encourage to pull positive samples (same age) closer within the local neighborhood and negative samples (different age) apart. In contrast, $\ell_{aco}$ pulls negative samples closer (different age) and positive samples (same age) apart. The goal here is to map all different ages within a very small neighborhood of each other, to deter the attacker from learning discriminative representations of age. Intuitively, this guides \textit{CBNS} to transform the released point cloud to introduce ambiguity in representations used by an attacker to correctly discriminate between ages, resulting in better privacy. In other words, the $\ell_{aco}$ forces to map the different ages to a single point in the embedding space.

The proxy attacker and proxy user have access to supervised data and attempt to minimize their losses $L_{util}$ and $L_{priv}$ respectively. CBNS is trained to minimize $L_{util}$ and maximize $L_{priv}$, simulating an implicit min-max optimization for these two components. Furthermore, CBNS also minimizes a soft-projection loss~\cite{lang2020samplenet} ($L_{sample}$) to improve stability of $f_{S}$ and ensure that the sampler is constrained to \textit{select} points from the input set (instead of interpolating). This overall objective can be summarized as:
\begin{equation}
\label{eq:total_loss}
    \min_ {\theta_S, \theta_D} \left[\max_{\theta_{A}} L_{priv}(\theta_S, \theta_D, \theta_{A}) + \lambda \min_{\theta_{S}, \theta_D, \theta_{U}} L_{util}(\theta_S, \theta_D, \theta_{U}) + 
    \min_{\theta_{S}} L_{sample}(\theta_S)
    \right]
\end{equation}

Here, $\lambda$ is a chosen hyperparameter to help regulate the trade-off between privacy and utility.

\vspace{-4mm}
\subsubsection{Inference:} A data owner with access to a point cloud dataset ($P, Y$) can use CBNS to generate the censored dataset ($\hat{P}, \hat{Y}$) and release it for use by untrusted parties. This released dataset can be used for either: i) training new models or ii) running inference using pre-trained models. This is possible only because the output space of the censoring mechanism is same as the input space. In other words, censoring a point cloud using CBNS also generates a point cloud. In contrast: i) most work for censoring images requires releasing neural activations which cannot be processed by arbitrary designed networks~\cite{singh2021disco,shredder-niloofar,osia} and ii) prior work for censoring point clouds releases line clouds~\cite{linecloud-Camera,linecloud-ImageBased} which, while useful for geometric tasks, are incompatible for off-the-shelf perception networks.

    





\section{Experiments}
\label{sec:expts}
In this section, we specify the datasets and baselines used, define the evaluation protocols and summarize implementation details for the results presented in this work. Details about the code are included in the appendix.

\noindent \textbf{Datasets:} \textbf{a) FaceScape}~\cite{yang2020facescape} consists of 16,940 textured 3D faces, captured from 938 subjects each with multiple categorical labels for age (100), gender (2) and expression (20). For our experiments, we sample 1024 3D points from the surface of each face mesh using Pytorch 3D~\cite{pytorch3d}. To simulate privacy-utility analysis, we use the expression as the task attribute (utility) and gender as the sensitive attribute (privacy). We choose this configuration because the default critical points for the two attributes overlap (need for noisy distorter) and are also distributed across the point cloud (need for invariant sampling), but human performance motivates that they are can be inferred independently. This provides a good benchmark for testing the efficacy of CBNS. \textbf{b) ModelNet}~\cite{modelnet} consists of 12,311 CAD-generated meshes across 40 categories (object types) of which 9,843 training and 2,468 testing data points. For our experiments, we uniformly sample 2048 3D points from the mesh surface and then project them onto a unit sphere. Since each input sample only has one attribute (object type), we adapt the strategy used by~\cite{max-entropy} to simulate our privacy-utility analysis. Specifically, to identify an additional attribute, we divide the 40 classes into two super-types: living and non-living. We anticipate living objects to have visually discriminative features instead of geometric shapes of non-living objects. For example, the task of classifying an object as living (\textit{person, plant}) or non-living (\textit{sofa, bed}) should not reveal any information about its underlying identity (\textit{person, plant, sofa, bed}). We use super-types as task attribute and object type as sensitive attribute. \\

\noindent \textbf{Baselines:}
\label{sec:baselines}
Prior work in censoring point clouds has largely focused on geometric tasks (image-based localization, SLAM, SfM, etc) via line clouds~\cite{linecloud-ImageBased,linecloud-Camera,linecloud-SFM,linecloud-SLAM}. However, our analysis (section~\ref{sec:discussion}) shows that line clouds are a weak baseline for perception tasks (classification, detection, etc). To ensure rigorous analysis, we define multiple baselines inspired by work in 2D vision that has focused on censoring images (and their activations) for perception tasks while mitigating attribute leakage. The baselines differ in the design of sampling and noisy distorter modules - which may be task-oriented (learned using data) or task-agnostic (deterministic). Our mechanism \textit{CBNS} is equivalent to the \textit{Oriented Sampling - Oriented Noise (OS-ON)} configuration. The baselines are summarized below, with more details in the appendix:
\begin{itemize}
    \item \textbf{Agnostic Sampling - Agnostic Noise (AS-AN):} Uses farthest point sampling~(FPS) with fixed gaussian distribution for noise. This is inspired from~\cite{fan2018image} which formalises differential privacy for images through Gaussian noise, without any learning. While~\cite{fan2018image} adds noise to images, we add it to a sampled point cloud obtained via FPS.
    \item \textbf{Agnostic Sampling - Oriented Noise (AS-ON):} Uses FPS and learns parameters of the gaussian distribution for noise (as in CBNS) using \textit{maximum likelihood} attacker training. This is inspired from~\cite{shredder-niloofar} which learn a noise distribution to obfuscate image activations. While~\cite{shredder-niloofar} adds noise to image activations, we add it to a sampled point cloud obtained via FPS.
    \item \textbf{Oriented Sampling - Agnostic Noise (OS-AN):} Uses differentiable point cloud sampling (as in CBNS) with fixed gaussian distribution for noise. This is inspired from~\cite{singh2021disco} which does channel pruning of image activations to remove sensitive information. While~\cite{singh2021disco} trains a DNN to prune neural activations, we train a DNN to sample point clouds~\cite{lang2020samplenet}.
\end{itemize}

\begin{figure}[t]
    \centering
    \begin{subfigure}[t]{0.47\linewidth}
            \centering
            \includegraphics[width=\linewidth]{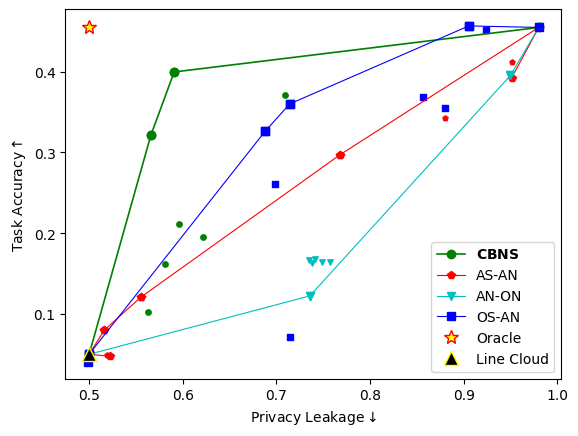}
            \caption{FaceScape \cite{yang2020facescape}}
    \end{subfigure}
    \begin{subfigure}[t]{0.47\linewidth}
            \centering
            \includegraphics[width=\linewidth]{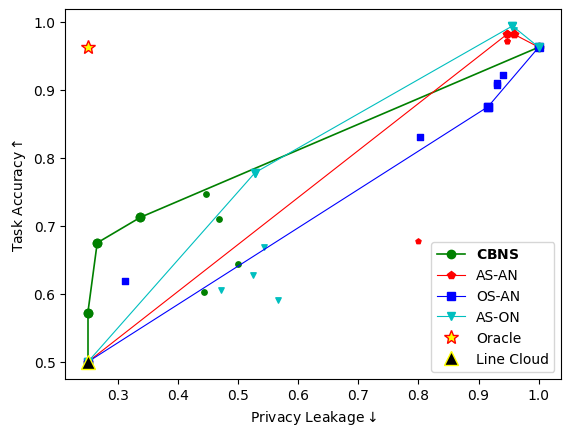}
            \caption{ModelNet \cite{modelnet}}
    \end{subfigure}
    \caption{\textbf{Privacy-Utility trade-off} comparison for different techniques by interpolating their best performing points. The oracle point refers to the best possible censoring mechanism. We note that the line cloud~\cite{linecloud-ImageBased,linecloud-SLAM} techniques do not yield any trade-off due to its incompatibility with perception tasks.}
    \label{fig:priv-util}
\end{figure}


\noindent \textbf{Evaluation Protocol:}
We evaluate different techniques by comparing the \textit{privacy-utility trade-off}. For each technique, utility is a measure of \textit{User}'s performance by training on the \textit{censored} dataset and privacy is of the \textit{Attacker}'s performance (as described in section~\ref{sec:formulation}). Specifically, we quantify information leakage from the dataset by comparing the performance of an attacker to correctly infer sensitive information from the censored dataset. For this analysis, we simulate a \textit{worst case} attacker that dynamically adapts to the privatization scheme. This adaptation is modeled using a pretrained attacker model that is fine-tuned on the \textit{censored} dataset and then evaluated on a censored test set. Inspired by~\cite{max-entropy}, we quantify privacy-utility trade-offs curves by different techniques using area under the pareto-optimal curve denoted as the normalized hypervolume (\textbf{NHV})~\cite{ishibuchi2018specify}. \textit{Higher NHV value indicates a better privacy-utility trade-off.}




\noindent \textbf{Implementation Details:}
Unless stated otherwise, we use 3D-point clouds ($d=3$) with the Invariant Sampler producing 64 points ($m=64$) and with PointNet~\cite{qi2017pointnet} as the backbone architecture for both $f_{U}$ and $f_{A}$. For FaceScape, $n=1024$ and the dataset split is 80\% training and 20\% testing examples. For ModelNet, $n=2048$ and we restrict experimentation to a smaller set of 4 classes: 2 living and 2 non-living to ensure ease of analysis and avoid data imbalance issues (ModelNet has 2 living and 38 non-living objects). All experiments are implemented using Pytorch and conducted on 2 TITAN-X GPUs. We will release both our code and dataset splits for reproducibility.

\section{Results}
\label{sec:results}

\begin{table}[t]
\centering
\begin{tabular}{l|c|c|c|c|c|c} 
\toprule
\multicolumn{1}{l|}{Method} & \multicolumn{3}{c|}{FaceScape} & \multicolumn{3}{c}{ModelNet} \\ 
\midrule
 & Privacy ($\downarrow$) & Utility ($\uparrow$) & NHV ($\uparrow$) & Privacy ($\downarrow$) & Utility ($\uparrow$) & NHV ($\uparrow$) \\ 
\midrule
No-privacy & 0.9515 & 0.4549 & - & 0.95 & 0.9625 & - \\ 
\midrule
Line Cloud~\cite{linecloud-ImageBased,linecloud-SLAM} & 0.5000 & 0.0500 & - & 0.2500 & 0.5000 & - \\
AS-AN~\cite{fan2018image} & 0.9511 & 0.4122 & 0.1524 & 0.9469 & 0.9612 & 0.6113 \\
AS-ON~\cite{shredder-niloofar} & 0.9391 & 0.3875 & 0.1250 & 0.5281 & 0.7781 & 0.6088 \\
OS-AN~\cite{singh2021disco} & 0.7143 & 0.3602 & 0.1439 & 0.3125 & 0.6187 & 0.6356 \\ 
\midrule
\textbf{CBNS (Ours)} & \textbf{0.5890} & \textbf{0.4013} & \textbf{0.1885} & \textbf{0.2657} & \textbf{0.6750} & \textbf{0.6492} \\
Oracle & 0.5000 & 0.4549 & - & 0.2500 & 0.9625 & - \\
\bottomrule
\end{tabular}
\caption{\textbf{Comparison for sensitive attribute leakage}. We compare our approach on sensitive attribute leakage with the existing works and baseline. CBNS outperforms Line Cloud~\cite{linecloud-ImageBased,linecloud-SLAM}, AS-AN~\cite{fan2018image}, AS-ON~\cite{shredder-niloofar}, OS-AN~\cite{singh2021disco} and achieves the best privacy-utility trade-off on FaceScape and ModelNet datasets. For the FaceScape, sensitive attribute is gender and task attribute is expression; and for the ModelNet, sensitive attribute is underlying object type and task attribute is  super-types (living or non-living).
\label{tab:baseline-compare}}
\end{table}

We report performance comparison with baselines on ModelNet and FaceScape, in Table~\ref{tab:baseline-compare}. For completeness, apart from the baselines defined in section~\ref{sec:baselines}, we also benchmark with two extreme scenarios: i) \textbf{No-privacy}: default case where the data is released without any censoring and, ii) \textbf{Oracle}: best possible case with some ideal censoring mechanism which \textit{does not exist}. Results show that \textbf{CBNS} significantly outperforms all baselines, on both datasets; as evident from higher NHV. \textit{First,} \textbf{CBNS} consistently provides the best privacy leakage - often very close to random chance. Specifically, privacy leakage with CBNS is 0.5890 for 2-way classification in FaceScape and 0.2657 for 4-way classification in ModelNet. \textit{Second,} the peak privacy-utility trade-off for \textbf{CBNS} is closest to the oracle, on both datasets. Specifically, when CBNS is used for data release on FaceScape, the \textit{User} achieves a utility of \textit{0.4013} (88\% of the oracle) while the \textit{Attacker} performance is close to random chance (0.5890). This corresponds to \textit{13\% less privacy leakage while also providing 4\% more utility} than the closest baseline (OS-AN). However, a higher fall in utility is observed on ModelNet, which can be attributed to the fact that the task and sensitive attributes are more strongly coupled, than in FaceScape. \textit{Finally,} these observations are corroborated visually by the privacy-utility trade-off curves in Figure~\ref{fig:priv-util} where CBNS has the highest area-under-curve (correlated with the hyper-volume).


\section{Discussion}
\label{sec:discussion}
We analyse the impact of key design choices for censoring by noisy sampling. Specifically, we study the characterization of both \textit{CBNS} modules: i) Invariant Sampler and ii) Noisy Distorter; and impact of the perception network. For completeness, we also analyse the viability of line clouds for perception tasks. For ease of exposition, we restrict the scope of this analysis to FaceScape dataset.
\begin{table}[t]
\centering
\begin{tabular}{l|c|c|c}
\toprule
Technique   & Privacy ($\downarrow$) & Utility ($\uparrow$) & NHV ($\uparrow$) \\
\midrule
AS-ON~\cite{shredder-niloofar} ($\ell_{cce}$)                   & 0.9297   & 0.3954  & 0.1322 \\
OS-AN~\cite{singh2021disco} ($\ell_{cce}$)               & 0.6942  & 0.3602  & 0.1439 \\ 
\midrule
CBNS ($\ell_{me}$)        & 0.6839  & 0.413   & 0.1596 \\ 
CBNS ($\ell_{cce}$)                & 0.5787  & 0.3638  & 0.1615 \\ 
\textbf{CBNS ($\ell_{a}$)} & \textbf{0.5707}  & \textbf{0.3997}  & \textbf{0.1885} \\ 
\bottomrule
\end{tabular}
\caption{\textbf{Design of Invariant Sampler}. Privacy-utility trade-off is influenced by whether transformation is learned, and how it is learned. For FaceScape, sensitive attribute is gender and task attribute is expression. $l_{cce}$, $l_{me}$ and $l_{a}$ denotes cross entropy loss, max-entropy loss and CBNS loss respectively.
\label{tab:sampling-matters}}
\vspace{-8mm}
\end{table}

\noindent \textbf{$-$ Design of Invariant Sampler:} 
We study the role of two design choices: i) learning a task-oriented sampler and ii) the attacker loss that is used to optimize the parameters of the learned sampler. Results are presented in Table~\ref{tab:sampling-matters}. Please note that we follow the baselines from section~\ref{sec:expts} and explicitly mention ($L_{priv}$) the attacker objective in parenthesis. Specifically, $l_{cce}$ is the cross-entropy loss, $l_{a}$ is our proposed loss (equation~\ref{eqn:attacker-loss}) and $l_{me}$ is a max-entropy loss used in ~\cite{max-entropy}. We observe the following: \textit{First}, using a learned task-oriented sampler reduces privacy-leakage by 35\% without any loss to utility (row 1 vs 4 and 5) in contrast to a task-agnostic sampler. \textit{Second,} using the proposed adversarial contrastive loss in $L_{priv}$ improves privacy-utility trade-off by increasing utility by 3\% without any additional privacy leakage (row 4 vs 5). \textit{Third,} using $l_{me}$ improves utility but with a significant increase in privacy-leakage; as evident from lower NHV. $l_{me}$ is successful for images~\cite{max-entropy} but fails to generalize to point clouds, which can be attributed to the irregularity in the data structure.

\noindent \textbf{$-$ Design of Noisy Distorter:} We study the role of three design choices: i) learning task-oriented noise, ii) the attacker loss used to optimize parameters of the learned noise ($l_{a}$ or $l_{cce}$) and iii) the granularity of the noise parameters (\textit{shared v.s pointwise}). \textit{Shared} implies that each point is distorted using noise from the same learned distribution (i.e. $z_{s} \in R^{1}$) and \textit{Pointwise} implies that each point is distorted from a unique independently learned distributions (i.e. $z_{s} \in R^{r}$). Results are presented in Table~\ref{tab:noise-ablation}. We follow the baselines from section~\ref{sec:expts} and define \textit{OS} (row 1) as additional baseline which only uses sampling (\textit{without} noisy deformation). We observe the following: \textit{First,} using noise task-agnostic (row 1 vs 2), or ii) shared task-oriented (row 1 vs 3, 4) does not provide benefit; and are infact worse than no noise baseline ($SO$). \textit{Second,} \textit{pointwise} noise distributions \textit{significantly} improves performance. (row 3 vs 5; 4 vs 6). This increase in NHV as well as peak privacy-utility trade-offs can be attributed to improved flexibility for adapting to characteristics of sensitive, task attributes and their relationship.. \textit{Third,} the objective function used for learning noise is also important where using $l_{aco}$ in $l_{a}$ improves privacy-utility trade-off (row 3 vs 4; row 5 vs 6).
\begin{table}[t]
\centering
\begin{tabular}{l|c|c|c}
\toprule
Technique   & Privacy ($\downarrow$) & Utility ($\uparrow$) & NHV ($\uparrow$)    \\
\midrule
OS ($\ell_{cce}$)              & 0.5787  & 0.3638  & 0.1615 \\ 
OS-AN~\cite{singh2021disco} ($\ell_{cce}$)              & 0.6942  & 0.3602  & 0.1439 \\ 
\midrule
CBNS (\textit{shared}, $\ell_{cce}$)         & 0.9051  & 0.4532  & 0.1586 \\ 
CBNS (\textit{shared}, $\ell_{a}$)    & 0.8600  & 0.4217  & 0.1625 \\ 
\midrule
CBNS (\textit{pointwise}, $\ell_{cce}$)      & 0.5689  & 0.4013  & 0.1530 \\ 
CBNS (\textit{pointwise}, $\ell_{a}$) & \textbf{0.5707}  & \textbf{0.3997}  & \textbf{0.1885} \\ 
\bottomrule
\end{tabular}
\caption{\textbf{Design of Noisy Distorter}. Privacy-utility trade-off is influenced by the learned noise and the granularity of the noise parameters (\textit{shared v.s pointwise}). For FaceScape, sensitive attribute is gender and task attribute is expression. $l_{cce}$ and $l_{a}$ denotes cross entropy loss and CBNS loss respectively.
\label{tab:noise-ablation}}
\vspace{-8mm}
\end{table}

\noindent \textbf{$-$ Impact of Perception Network:} In our threat model, the \textit{Owner} releases dataset for post-hoc access by the \textit{User} and \textit{Attacker}. Hence, the censoring mechanism should be independent of the type of perception networks used by these entities for downstream tasks. We analyse this sensitivity by comparing two different proxy attacker and user networks. Specifically, we use DGCNN~\cite{dgcnn} which is a recent state-of-the-art network with higher capacity and distinct saliency properties than PointNet~\cite{pc-saliency}. Results are presented in Table~\ref{tab:perception-and-lines}. We observe the following: \textit{First,} increasing capacity of proxy networks further improves learning of CBNS as evident from better privacy-utility trade-offs (row 1 vs 2; row 3 vs 4). For instance, when CBNS is trained with DGCNN (as against PointNet), the censored dataset provides a better utility of 0.4100 (vs 0.3997) while also significantly reducing privacy leakage to 0.4848 (vs 0.5707). \textit{Second,} importantly, we see that noisy sampling is independent of the downstream network and can generalize to multiple perception networks. Specifically, this is very encouraging since CBNS can concurrently mitigate stronger \textit{attackers} from leaking information by improving the utility of \textit{users} with the stronger perception backbones.
\begin{table}[t]
\centering
\begin{tabular}{l|c|c|c|c}
\toprule
Technique              & Backbone      & Privacy ($\downarrow$) & Utility ($\uparrow$) & NHV ($\uparrow$) \\
\midrule
\multirow{2}{*}{OS-AN~\cite{singh2021disco}} & PointNet & 0.6942  & 0.3602  &  0.1439   \\
                       & DGCNN    & 0.7056  & 0.4718  &  0.2136  \\
\midrule
\multirow{2}{*}{\textbf{CBNS (Ours)}}  & PointNet & 0.5707  & 0.3997  & 0.1885  \\
                       & DGCNN    & \textbf{0.4848}  & \textbf{0.4100}  &  \textbf{0.2361}   \\
\midrule
\midrule
\multirow{2}{*}{Line Cloud~\cite{linecloud-ImageBased,linecloud-SLAM}}  & PointNet & 0.5000  & 0.0500  & -    \\
                       & DGCNN    & 0.5000  & 0.0500  &  -   \\
                   
\midrule
\end{tabular}
\caption{\textbf{Impact of Perception Network} and \textbf{Incompatibility of Line Clouds}. For FaceScape, sensitive attribute is gender and task attribute is expression.
CBNS is invariant to the type of attacker network. Using stronger perception network (DGCNN) further improves performance over PointNet and helps achieve near optimal trade-off with our proposed CBNS. Resampling line clouds provides poor (random chance) privacy-utility trade-off. \label{tab:perception-and-lines}}
\vspace{-8mm}
\end{table}

\noindent \textbf{$-$ Incompatibility of Line Clouds:} We posit that line clouds are an incompatible baseline for perception tasks because i) they destroy semantic structure of the input point cloud which is essential for perception (see visualizations in~\cite{linecloud-ImageBased}), and ii) any off-the-shelf perception network: used by both \textit{User} and \textit{Attacker} cannot train on line clouds. To benchmark the performance of line clouds, we generate point clouds by re-sampling line clouds and evaluate performance on our perception queries. Results in Table~\ref{tab:perception-and-lines} show that: i) re-sampling line clouds provides extremely poor utility (random chance), and ii) the privacy-utility trade-off cannot be tuned (hence no NHV). Specifically, we observe that resampled line cloud obtain privacy leakage of 0.5000 (for 2-way classification) and utility of 0.05 (on a 20-way classification). Finally, we acknowledge recent work has attacked line clouds to reconstruct point clouds~\cite{chelani2021privacypreserving} but note that this is equivalent to our No-Privacy baseline which can improve utility but requires mechanisms like noisy sampling to provide privacy.

\section{Related Work}

\noindent \textbf{Private Imaging}.
A majority of the existing works in privately sharing data focus on identifiability and anonymization~\cite{samarati1998protecting,dwork2006calibrating,wang2020comprehensive}. 
In contrast to this line of work, we focus on protecting sensitive attributes. Among the techniques that focus on protecting sensitive attributes~\cite{shredder-niloofar,singh2021disco,xiao2020adversarial,osia,liu2020datamix}, their tasks are typically limited to image datasets. 
More recently, privacy for 3D point clouds has emerged with a focus on geometric queries protecting privacy by releasing line clouds~\cite{linecloud-ImageBased,linecloud-SFM}. However, 
To the best of our knowledge, this is the first work in protecting sensitive information leakage for perception tasks in point clouds. Adjacent to research in privately sharing data, privately sharing ML model~\cite{mcmahan2016federated,GUPTA20181,dpsgd,dpgan,jordon2018pate} has received interest recently. 
However, unlike protecting sensitive attributes, model sharing aims to protect the identifiability of training data.

\noindent \textbf{Learning on Point Clouds}.
Recent advances in deep learning (DL) have allowed to learn directly on raw point clouds; enabling use in diverse perception tasks such as classification, semantic segmentation, object detection, registration etc. Various DL architectures have been proposed starting with PointNet~\cite{qi2017pointnet}, PointNet++~\cite{qi2017pointnet++} and follow-up works in ~\cite{liu2020closer,hu2020randla,yan2020pointasnl,bytyqi2020local,xu2020grid,xiang2021walk,lin2021point2skeleton,wu2019pointconv,li2018pointcnn} that improve the performance over a given task by capturing task-oriented representations.
Zheng \textit{et al.}~\cite{pc-saliency} observe that saliency of the point cloud networks is localized and network rely on a small subset of the signal for the task. This observation has led to extensive work in privacy and security~\cite{adv-attack-defense,yang2021adversarial,liu2019extending,adv-attack} utilizing the localized saliency used to design adversarial attack (and defence) mechanisms on the trained models. We note that our setting significantly differs from adversarial attack work since we \textit{protect the dataset} that can be used to train arbitrary models while adversarial methods focus on \textit{attacking/protecting} the robustness of model predictions. 

\noindent \textbf{Sampling of Point Clouds}.
Processing point clouds can be computationally intensive making sampling a popular pre-processing step to alleviate this challenge. Classical methods such as random sampling and FPS~\cite{qi2017pointnet++,li2018pointcnn} are task-agnostic and deterministic algorithms for sampling point sets. However, not utilizing task knowledge when sampling hinders performance. Recent techniques~\cite{Dovrat2019LearningTS,lang2020samplenet} introduce task-oriented mechanisms for sampling through differentiable approximations. The focus is to improve compute efficiency while preserving the entire signal in the sampled subset. In contrast, our goal is to censor sensitive information during the sampling process. We build upon prior work to introduce a task-oriented point-cloud sampler that censors sensitive information.

\noindent \textbf{Noisy Sampling for Censoring}. While not motivated for point clouds, similar intuition has been used for tabular datasets for private coresets~\cite{gupta2010differentially,feldman2009private} combines subset (coreset) selection and differentially-private noise to achieve good privacy utility trade-off. Our work is different because: i) our queries involve neural networks so computing sensitivity for DP-noise is infeasible, ii) we only want to protect sensitive attribute. We empirically validate privacy-utility trade-off using benchmark metrics, as described in section~\ref{sec:expts} and present results in section~\ref{sec:results}.

\section{Conclusion}
\vspace{-2mm}
This focus of this paper is to censor point clouds to provide utility for perception tasks while mitigating attribute leakage attacks. The key motivating insight is to leverage the localized saliency of perception tasks on point clouds to provide good privacy-utility trade-offs. We achieve this through our mechanism called censoring by noisy sampling (\textit{CBNS}), which is composed of two modules: i) Invariant Sampling - a differentiable point-cloud sampler which learns to remove points invariant to utility and ii) Noise Distorter - which learns to distort sampled points to decouple the sensitive information from utility, and mitigate privacy leakage. We validate the effectiveness of CBNS through extensive comparisons with state-of-the-art baselines and sensitivity analyses of key design choices. Results show that CBNS achieves superior privacy-utility trade-offs.

\bibliographystyle{splncs}
\bibliography{main}
\end{document}